\documentclass{ifacconf}                                                          

\counterwithin*{section}{part}

\usepackage{amsmath} 
\usepackage{amssymb}  
\usepackage{import}
\usepackage{graphicx}
\usepackage{tikz}
\usepackage{pgfplots}
\usepackage{subfigure}

\usepackage{comment}
\usepackage{cases} 

\makeatletter
\let\old@ssect\@ssect 
\makeatother

\usepackage{hyperref}

\makeatletter
\def\@ssect#1#2#3#4#5#6{%
  \NR@gettitle{#6}
  \old@ssect{#1}{#2}{#3}{#4}{#5}{#6}
}
\makeatother

\makeatletter
\long\def\@makefigurecaption#1#2{\@figurecaptionsize
  \vskip \@overcaptionskip
  \setbox\@tempboxa\hbox{#1. #2}
  \ifdim \wd\@tempboxa >\hsize              
    \unhbox\@tempboxa\par                   
  \else                                     
    \global \@minipagefalse
    \hbox to\hsize{\hfil\unhbox\@tempboxa\hfil}
  \fi}
\makeatletter

\usepackage{cleveref}
\usepackage{multirow}
\usepackage{algorithm}
\usepackage[noend]{algpseudocode}
\usepackage{relsize}
\usepackage{xcolor}
\usepackage{natbib}        
\usepackage{tabularx,booktabs}
\usepackage{wrapfig}
\newcolumntype{C}{>{\centering\arraybackslash}X} 
\algnewcommand{\algorithmicor}{\textbf{ or }}
\algnewcommand{\OR}{\algorithmicor}
\newtheorem{definition}{Definition}
\newtheorem{theorem}{Theorem}

\newtheorem{lemma}{Lemma}
\newtheorem{hypothesis}{Hypothesis}

\DeclareMathOperator*{\argmin}{arg\,min}
\crefname{figure}{Fig.}{Fig.}
\Crefname{figure}{Figure}{Figures}
\Crefname{equation}{Equation}{Equations}
\crefname{theorem}{Theorem}{Theorems}
\crefname{section}{Section}{Sections}
\crefname{hypothesis}{Hypothesis}{Hypotheses}
\crefname{proposition}{Proposition}{Propositions}
\definecolor{mycolor1}{rgb}{0,0.5,1}%
\definecolor{mycolor2}{rgb}{0.4,0.9,0.6}%
\definecolor{mycolor3}{rgb}{0.6,0.7,0.4}%
\definecolor{mycolor4}{rgb}{0.8,0.5,0.2}%
\definecolor{mycolor5}{rgb}{0.9,0.4,0.1}%

\begin{document}

\begin{frontmatter}

\title{Safe Adaptation\\ with Multiplicative Uncertainties\\ Using Robust Safe Set Algorithm}

\author[First]{Charles Noren} 
\author[First]{Weiye Zhao} 
\author[First]{Changliu Liu}

\address[First]{Carnegie Mellon University, Pittsburgh, PA 15213, USA \\(e-mails: cnoren, weiyezha, cliu6@andrew.cmu.edu).}
\thanks{This work is in part supported by Amazon Research Award.}

\begin{abstract}
Maintaining safety under adaptation has long been considered to be an important capability for autonomous systems. As these systems estimate and change the ego-model of the system dynamics, questions regarding how to develop safety guarantees for such systems continue to be of interest. We propose a novel robust safe control methodology that uses set-based safety constraints to make a robotic system with dynamical uncertainties safely adapt and operate in its environment.  The method consists of designing a scalar energy function (safety index) for an adaptive system with parametric uncertainty and an optimization-based approach for control synthesis. Simulation studies on a two-link manipulator are conducted and the results demonstrate the effectiveness of our proposed method in terms of generating provably safe control for adaptive systems with parametric uncertainty.
\end{abstract}


\end{frontmatter}

\section{INTRODUCTION } \label{sec: introduction}
The ability to adapt is important for autonomous systems in many real-world scenarios, such as when an autonomous vehicle enters new road and traffic conditions, or when a robotic system suffers component failures (e.g., a broken motor). The purpose of adaptation is to optimize the performance of the autonomous system online in new scenarios without offline retraining or redesign. Since the adaptation process involves online interactions with uncertain environments, it must be efficient (i.e., converging fast to the ground truth) and safe (i.e., maintaining the system state in a safe set). Energy function based safe control approaches for known dynamical systems have been widely applied to ensure safety as discussed in \cite{Ames_2019} and \cite{TC_SCAUEF}. On the other hand, robust control and adaptive control have been applied for parametric identification in uncertain environments, as discussed in \cite{JJS_book}. However, techniques for ensuring safe system behavior, i.e., those beyond stability or performance-based metrics, under parametric uncertainty through model-based safe control has not been extensively studied. This paper develops an approach to ensure safe and efficient adaptation under parametric uncertainty.

Energy function based safe control methods consist of two components. The first component is to define a scalar function over the system state to measure the ``energy'' of a system, in other words, how far the system is away from a target safe set. By making the system dissipate energy if the safe set is not achieved or preserve energy if the safe set is achieved, we can ensure forward invariance and global attractiveness of the safe set.
These energy functions often take a variety of names, including safety index, potential function, barrier function, and most commonly: Lyapunov function, as discussed in \cite{TC_SCAUEF}. There exists a fundamental equivalency between the objective of the approaches, but the specific control response taken differs. The second component is the actual control synthesis to realize the forward invariance of the safe set, where the system model is explicitly considered. 

However, uncertainty in the system parameters could drive an incorrect system response and break the forward invariance guarantees. Methods to handle additive uncertainties in the dynamics, i.e., uncertainties that can be wholly accrued into the drift term ($f(x)$ in \eqref{eqss}) are discussed in \cite{L_SEA}, \cite{TA_ASCBF}. The core idea is to increase the safety margin in the control synthesis with respect to the additive uncertainties. 

Noticeably, these methods do not address multiplicative uncertainties in the dynamics which modify the control term ($g(x)$ in \eqref{eqss}). With multiplicative uncertainties, the direct effect of certain control inputs is less predictable. Existing approaches which solely increase the tolerance on the safety margin will not guarantee the set invariance of the safe set in this case, and thus new methods need to be developed. It is worth noting that the method in \cite{pmlr-v120-taylor20a} addresses uncertainty in both terms by learning the residual lie derivatives and then using this learned model to satisfy the energy function constraint. They assume any valid energy function for the learned dynamic system is valid for the real dynamics system, but no theoretical proof is provided for this assumption. We provide a proof of a safety guarantee given specific conditions on the safety controller and the uncertainties on the parameters of $g(x)$.

\color{black}


This paper proposes the Robust Safe Set Algorithm (RSSA) to ensure the safe adaptation of uncertain systems with parametric uncertainties that propagates into both defining functions of the system dynamics: e.g., $f(x)$ and $g(x)$ in \eqref{eqss}. RSSA can be applied in a supervisory loop for any model-based adaptive control algorithm as long as the underlying adaptive control algorithm estimates the unknown parameters and the associated uncertainties. To ensure safety, RSSA modifies both components of traditional energy function based safe control approaches in order to enable the control system to directly account for the parametric uncertainties. The first modification is through the design of a new extension to the safety index, named a Composite Safety Index, which in its presented form penalizes deviations from the expected value of the system parameters to encourage a conservative safety behavior. The second modification  is to extend the Safe Set Algorithm (SSA) in \cite{LT_CIASS}, which does not explicitly account for any uncertainty, and the Safe Exploration Algorithm (SEA) in \cite{L_SEA}, which only accounts for additive uncertainty, to account for multiplicative uncertainties in the control synthesis. 
The algorithm is proved to ensure forward invariance of the safe set when the uncertainty is bounded and sufficiently small. The theoretical contributions presented are then numerically validated in simulation.



\section{PROBLEM DESCRIPTION } \label{sec: probdesc_sec2}
Consider a state space, $\mathcal{X} \subset \mathbb{R}^n$, a control space, $U \subset \mathbb{R}^m$, and a control-affine dynamic system, given by:
\begin{equation}
    \dot{x} = f(x) + g(x)u, \label{eqss}
\end{equation}
where $f(x)$ and $g(x)$ are Lipschitz continuous. For all dynamic systems, an equation in the form \eqref{eqss} may always be formed via dynamic extension according to \cite{isidori2013nonlinear}, thus \eqref{eqss} reflects a given general form of a dynamical system. Consider a modification to \eqref{eqss}, where both $f(x)$ and $g(x)$ are now also dependent on an unknown bounded parameter $\xi \in \mathbb{R}^p$, an element of parameter space $\mathcal{P}$, the system estimate of which is $\hat{\xi}$, and the true value of the parameter is $\xi_T$. Furthermore, consider that the initial parameter estimate and the true parameter value are contained within the parameter bounds, $\xi, \xi_T \in [\underline{\xi}, \bar{\xi}]$. Rewriting \eqref{eqss} to include this new parameter, the dynamic system equation becomes:
\begin{equation}
    \dot{x} = f(x, \xi) + g(x, \xi)u. \label{eqssAC}
\end{equation}
We assume there is an adaptive control algorithm that estimates $\hat\xi$ and the associated uncertainty $\Xi_\xi$ and then synthesizes the control $u$ to achieve the control objective of the system (e.g., trajectory tracking, goal reaching, regulation, etc). We want to ensure the adaptation is done safely, i.e., the system trajectory will always lie in a specific set: $\mathcal{X}_S\subset\mathcal{X}$, called the safe set. 
For simplicity, this paper assumes that the control $u$ is unbounded. The case that $u$ is bounded will be left for future work.

\section{REVIEW OF SAFE CONTROL} \label{sec: roscp_sec3}
The design objective for safe control is to maintain the system state in the safe set $\mathcal{X}_S$. 
The safe set is conventionally defined such that $\mathcal{X}_S$ is a 0-sublevel set of a continuously differentiable function $\phi_0: \mathcal{X} \rightarrow \mathbb{R}$, where $\phi_0(x)$ is known as the initial safety index. The initial safety index maps the subset of ``unsafe" states in the state space to positive real values and ``safe" states to zero or negative real values. 
If a control law is selected such that the closed-loop system dynamics always satisfy the following conditions, the safe set will be forward invariant and globally attractive according to \cite{rauch1978qualitative}:
\begin{equation}
    \dot{\phi}_0(x) \leq -\eta, \text{ } \forall \text{ } x \in \{ x \text{ }: \text{ } \phi_0(x) \geq 0\}, \label{eqCC}
\end{equation}
for some $\eta>0$. We say the system is safe if the safe set is forward invariant (never leaves the safe set) and globally attractive (always goes back to the safe set). 

However, the above conditions may not be physically realizable if the control input $u$ cannot directly affect $\dot{\phi}_0$, mathematically stated as:  $\frac{\partial\dot{\phi}_0}{\partial u} = 0$. Therefore, to realize safe control, the designer must select a safety index,  $\phi: \mathcal{X} \rightarrow \mathbb{R}$, which is not necessarily the same as the initial safety index, but whose time derivative can be directly influenced by the control so that the following conditions can be directly enforced:
\begin{equation}
    \dot\phi(x)\leq -\eta, \forall \text{ } x\in\{x:\phi(x)\geq 0\}.\label{eq: phi dot}
\end{equation}
In the following discussion, denote the 0-sublevel set of $\phi$ as $\mathcal{L}(\phi):=\{x:\phi(x)\leq 0\}$. Define $X(\phi)$ as the set of reachable states if the system starts from $\mathcal{X}_S\cap \mathcal{L}(\phi)$ and evolves under the constraint \eqref{eq: phi dot} without considering the limitations posed by the actual system dynamics \eqref{eqss}.\footnote{In other words, we can regard $X(\phi)$ as the reachable set under an artificial system dynamics $\dot x = u$ and the constraint \eqref{eq: phi dot}.} $X(\phi)$ is called the reachable set under $\phi$. By definition, $X(\phi)\subseteq \mathcal{L}(\phi)$.
To ensure safety, the safety index $\phi$ should be designed as a smooth function that satisfies \Cref{def: sufficient}. 
\begin{definition}[Safety Index Applicability Conditions]\label{def: sufficient}
The safety index is valid if: 1) the control input can influence its time derivative, i.e., $\frac{\partial \dot{\phi}}{\partial u} \neq 0$; and 2) the reachable set under $\phi$ is a subset of the safe set, mathematically: $X(\phi)\subseteq \mathcal{X}_S$.
\end{definition}
The first condition ensures that the control can influence the safety index (controllability). The second condition requires that the designed safety index constrains the system in the same way as the initial safety index (similarity). The work in \cite{LT_CIASS} discussed approaches to construct $\phi$ from $\phi_0$ to meet the two conditions. 
 If these conditions hold, then we can ensure the safety of the dynamic system \eqref{eqss} by choosing a control law that satisfies \eqref{eq: phi dot}. The constraint in \eqref{eq: phi dot} can be converted to a constraint on the control by incorporating the system dynamics \eqref{eqss}, since
$
    \dot \phi(x) = L_f\phi + L_g\phi ~ u
    $ 
where $L_f \phi = \frac{\partial \phi}{\partial x} f(x)$ and $L_g\phi=\frac{\partial \phi}{\partial x} g(x)$ are Lie derivatives. We define the set of safe control as
\begin{equation}\label{eq: control constraint}
    \mathcal{U}_S(x):=\{u\mid L_f\phi + L_{g}{\phi} \cdot u \leq -\eta, \text{ when }\phi(x)\geq 0\}.
\end{equation}
For systems without control limits, the set of safe control is non-empty since $L_g\phi$ is non-zero according to the controllability condition. The SSA projects a nominal control input $u^r$ to the set of safe control through the following quadratic program:
\begin{equation}
    \min_{u\in\mathcal{U}_S} \|u-u^r\|^2.
\end{equation}
The SEA considers the additive uncertainty $f\in \Sigma_f(x)$ and projects $u^r$ to a robust version of \eqref{eq: control constraint}.
\begin{subequations}
\begin{align}
    \min_u ~& \|u-u^r\|^2,\\
    \text{s.t. }& \max_{f\in\Sigma_f(x)}L_f\phi + L_{g}{\phi} \cdot u \leq -\eta, \text{ when }\phi(x)\geq 0.
\end{align}
\end{subequations}
How to address the multiplicative uncertainty, i.e., the uncertainty in $g(x)$, is the primary focus of this paper. 

\section{ROBUST SAFE SET ALGORITHM}\label{sec: method}

This section introduces the Robust Safe Set Algorithm (RSSA). We first discuss an approach to incorporate uncertainty estimation in the safety index, then introduce the approach to compute a robust set of safe control under parametric uncertainties.  
\subsection{Composite Safety Index Design}
\begin{wrapfigure}{r}{0.25\textwidth}
\vspace{-10pt}
    \centering
    \includegraphics[width=0.23\textwidth]{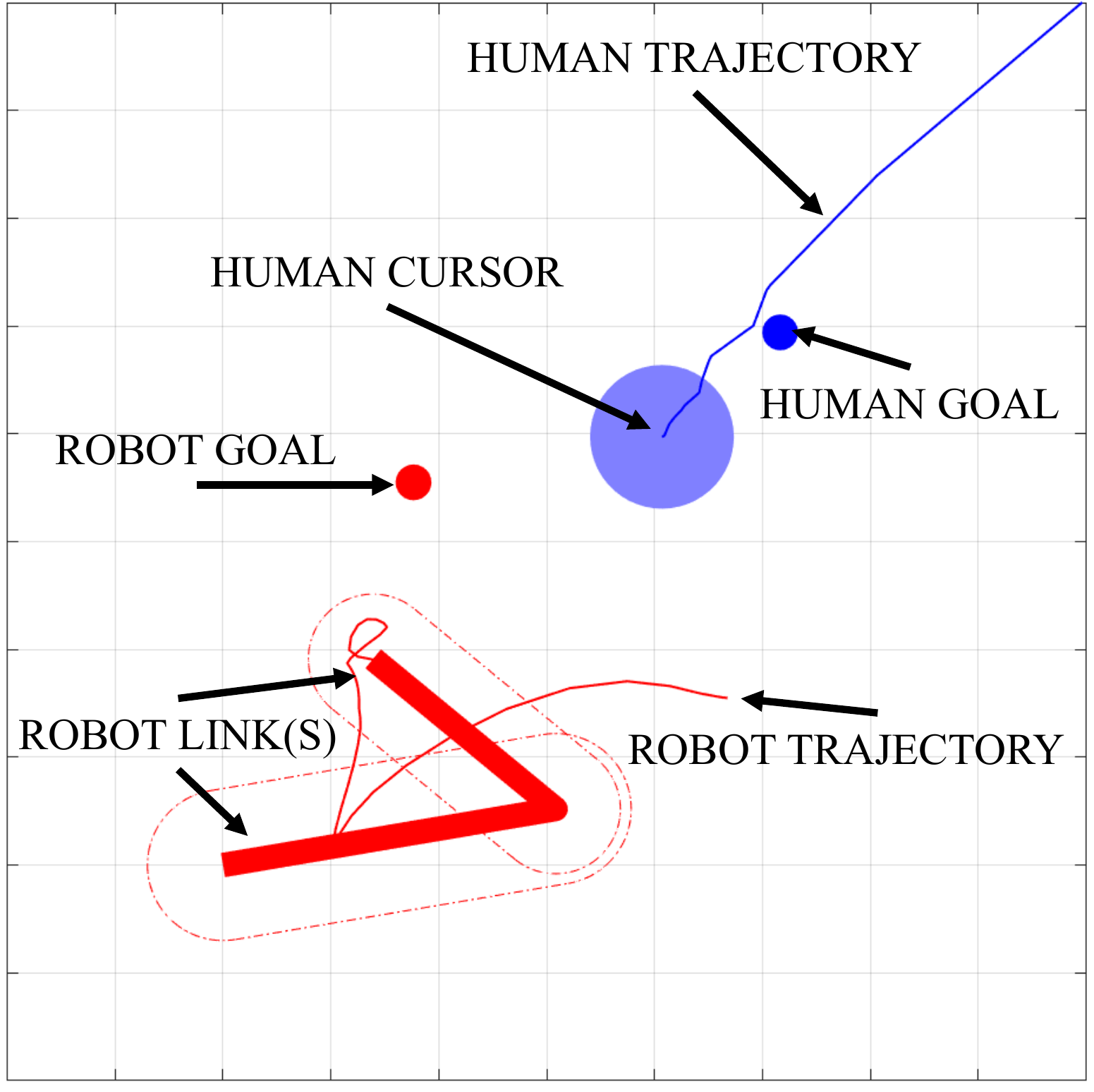}
    \caption{Numerical study setup.}
    \label{fig:setup}
\vspace{-10pt}
\end{wrapfigure}
Suppose we have a smooth safety index $\phi:\mathcal{X}\rightarrow \mathbb{R}$ that satisfies \Cref{def: sufficient} and does not include any term associated with the parametric uncertainty in $\xi$. While parametric uncertainty in the dynamics makes the satisfaction of \eqref{eq: phi dot} difficult, we extend the safety index $\phi(x)$ by an additive term $\phi_{\alpha}(\hat\xi,\Xi_{\xi})$ that penalizes parametric uncertainty to improve the safe control performance. However, $\phi_{\alpha}$ may not be arbitrarily selected as the new safety index with the additive term must continue to satisfy \Cref{def: sufficient} in order to enforce the original safety criterion. As $\phi$ already satisfies the controllability condition (hence $\phi+\phi_{\alpha}$ also satisfies the same condition), we only need to ensure:
\begin{equation}\label{eq: composite safety index}
    X(\phi + \phi_{\alpha}) \subseteq \mathcal{X}_S ,
\end{equation}

\begin{lemma}[Composite Safety Index]\label{thm:esi}
The condition \eqref{eq: composite safety index} holds if $\phi$ satisfies \Cref{def: sufficient} and $\phi_{\alpha}$ is a smooth function in time that is positive semi-definite.
\end{lemma}
The proof is shown in Appendix \ref{PCSID}. \Cref{thm:esi} provides conditions for the Composite Safety Index to satisfy \Cref{def: sufficient}. 
The conclusion is general for additive terms, but is specifically used here to introduce information about the system parametric uncertainty into the safety index. We call the specific form of safety index which includes parameter information a Robust Safety Index (RSI). The RSI is a function on the state, the parameter estimate, and the uncertainty. We may define $\phi_\alpha$ as the norm of $\Xi_\xi$ or any other form that ensures positive semi-definiteness. 

\subsection{Robust Safe Set Algorithm}

As the SEA is able to handle uncertainty in $f(x)$, this section will introduce a method to handle uncertainty in $g(x)$. 
For a given and known structure of $g(x)$, the parameter uncertainty in ${\xi}$ defines a family of functions, denoted as $\Sigma_g(x)$. 
To ensure safety, we need to choose the control signal so that \eqref{eq: phi dot} is satisfied for all possible $g(x)$. Note that with the Composite Safety Index, the constraint in \eqref{eq: phi dot} needs to be modified to be $\dot\phi + \dot\phi_\alpha \leq -\eta$ whenever $\phi+\phi_\alpha\geq 0$. Nonetheless, we can define $\eta(t) = \eta + \dot\phi_\alpha$ when $\phi+\phi_\alpha\geq 0$ and $\eta(t)=-\infty$ when $\phi+\phi_\alpha< 0$. Then the constraint reduces to $\dot\phi\leq-\eta(t)$. In the following discussion, we will use this simplified constraint. 
The robust set of safe control under multiplicative uncertainty is defined as:
\begin{align} \label{eq: robust safe set in g}
    \bar{\mathcal{U}}_S(x, &\Sigma_g(x)) =\bigcap_{g\in\Sigma_g(x)}\mathcal{U}_S(x)\\
    &= \{u\mid  L_f\phi + L_g\phi\cdot u\leq -\eta(t),\forall g(x)\in\Sigma_g(x) \nonumber\}.
\end{align}
For any arbitrary system, \eqref{eq: robust safe set in g} is not guaranteed to be non-empty. Consider: there may be $g_1(x),g_2(x)\in\Sigma_g$ such that $L_{g_1}\phi = - L_{g_2}\phi$. In this case, if $L_f\phi +\eta(t)>0$, then there is no control that can make \eqref{eq: control constraint} hold for all $g(x)\in\Sigma_g(x)$. However, if there exists $\alpha\in(0,1]$ and $\beta >0$ such that
\begin{align}\label{eq: uncertainty requirement for g}
    &L_{g_1}\phi \cdot L_{g_2}\phi \geq \alpha \|L_{g_1}\phi\|\|L_{g_2}\phi\|, \forall g_1(x),g_2(x)\in\Sigma_g(x), \nonumber\\
    &\|L_g\phi\|\geq \beta,\forall g(x)\in \Sigma_g(x),
\end{align}
then the set in \eqref{eq: robust safe set in g} is non-empty as shown in \cref{lemma: existence of robust safe control}.
The above condition \eqref{eq: uncertainty requirement for g} can be satisfied either when we have small initial uncertainty range $\Sigma_g(x)$ or when the adaptation algorithm will attenuate the uncertainty. 
\begin{lemma}[Feasibility of robust safe control]\label{lemma: existence of robust safe control}
When \eqref{eq: uncertainty requirement for g} holds, the set in \eqref{eq: robust safe set in g} is non-empty.
\end{lemma}
The proof is shown in \Cref{appendix: lemma 2 proof}. Within the robust set of safe control, we want to find an ``optimal'' safe control signal such that we can ensure safety with minimum control effort.
\begin{subequations}\label{eq: find g*}
\begin{align}
    \min_{u}~ & \|u\| \\
    \text{s.t. } & L_f\phi + L_g\phi\cdot u \leq -\eta(t), \forall g(x)\in\Sigma_g(x).
\end{align}
\end{subequations}
\begin{lemma}[Minimum effort robust safe control]\label{lemma: minimum effort}
When we have $L_f\phi+\eta(t) \leq 0$, the optimal solution of \eqref{eq: find g*} is $u=0$. When $L_f\phi +\eta(t)>0$, the optimal solution of \eqref{eq: find g*} is
\begin{align}\label{rssa}
    u = -{\frac{L_f\phi +\eta(t)}{\alpha^*}} L_{g^*}\phi, \alpha^* = \min_{g(x)\in\Sigma_g(x)} {L_{g^*}\phi \cdot L_g\phi},
\end{align}
where $g^*(x)$ is the solution of
\begin{align}\label{eq: find g* equivalent problem}
    \max_{g^*(x)\in\Sigma_g(x)}\min_{g(x)\in\Sigma_g(x)}\frac{L_{g^*}\phi \cdot L_g\phi}{\|L_{g^*}\phi\|}.
\end{align}
\end{lemma}

The proof is shown in \Cref{appx3}. In the case that we need to consider both the additive and multiplicative uncertainties, we can select the control:
\begin{align} \label{eq: final_law}
    u = -{\frac{\max_{f(x)\in\Sigma_f(x)}L_f\phi +\eta(t)}{\alpha^*}} L_{g^*}\phi.
\end{align}
This solution may not be optimal (i.e., $\|u\|$ may not be minimal) as it ignores the correlation between $\Sigma_f(x)$ and $\Sigma_g(x)$; but is valid. It is easy to verify that \eqref{eq: final_law} lies in the robust set of safe control that considers uncertainty in both $f(x)$ and $g(x)$.

\begin{algorithm}
\caption{Robust Safe Set Algorithm}\label{alg:RSSA}
\begin{algorithmic}[1]
\Procedure{RSSA}{$\eta$, $\phi(x)$, $\phi_\alpha$, ADAPT} 
\State Compute reference control $u^r$ and parameter estimate $(\hat\xi, \Xi_\xi)$ using ADAPT
\State Set $u=u^r$
\If {$\phi +\phi_\alpha > 0$}
\State Compute $\Sigma_g(x)$ using $(\hat\xi, \Xi_\xi)$.
\State Find $L_{g^*}\phi$ via \eqref{eq: find g* equivalent problem}
\If{$u^r \not \in \bar{\mathcal{U}}_S(x, \Sigma_g(x))$}
\State Compute robust safe control ($u$) via \eqref{eq: final_law}.
\EndIf
\EndIf
\State Return control $u$.
\EndProcedure
\end{algorithmic}
\end{algorithm}


Our algorithm is summarized in \Cref{alg:RSSA}, which efficiently finds the safe control under system parametric uncertainty when the nominal control generated by a nominal adaptive controller (denoted as ADAPT) is unsafe. We summarize the main result of this paper below. 
\begin{theorem}[Safety guarantee of \Cref{alg:RSSA}]\label{thm:main}
Consider a system with the dynamics $\dot x = f(x)+g(x)u$ where $g(x)$ is uncertain. If 1) $\Sigma_g$ correctly bounds the parametric uncertainty in $g(x)$, 2) \eqref{eq: uncertainty requirement for g} holds, 3) $\phi$ satisfies \Cref{def: sufficient}, and 4) $\phi_\alpha$ is smooth and positive semi-definite, then \Cref{alg:RSSA} ensures forward invariance of the safe set with minimum effort.

The proof of \Cref{thm:main} is shown in \Cref{the1}.
\end{theorem}

\section{SAFE ADAPTATION FOR ROBOTIC ARM} \label{sec: results}
\begin{figure*}[pt!]
  \centering
  {   
     \subfigure[\tiny M1,$\,$k$=200$\normalsize]{\label{fig:t2_ghat_phi_200graph}
       \includegraphics[width=0.725in]{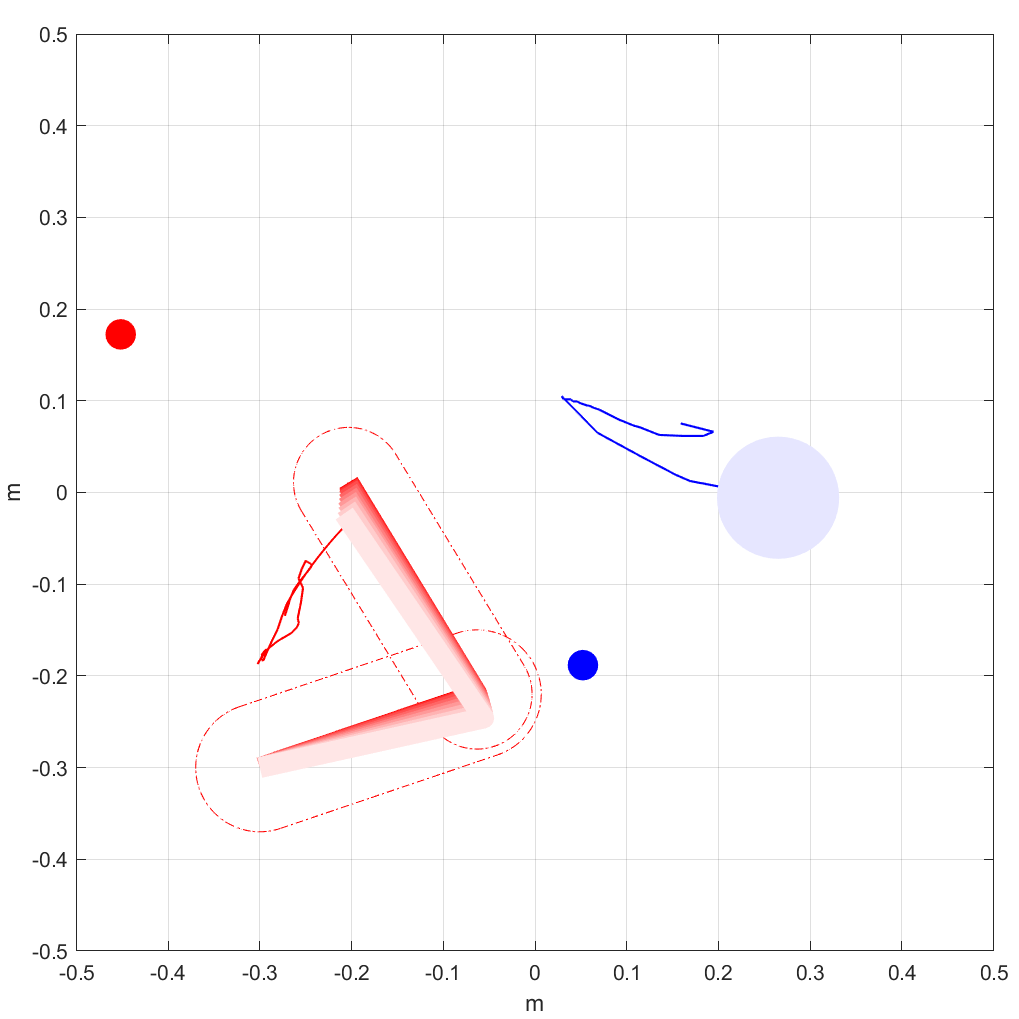}}   
     \;
     \subfigure[\tiny M1,$\,$k$=750$\normalsize]{\label{fig:t2_ghat_phi_750graph}%
         \includegraphics[width=0.725in]{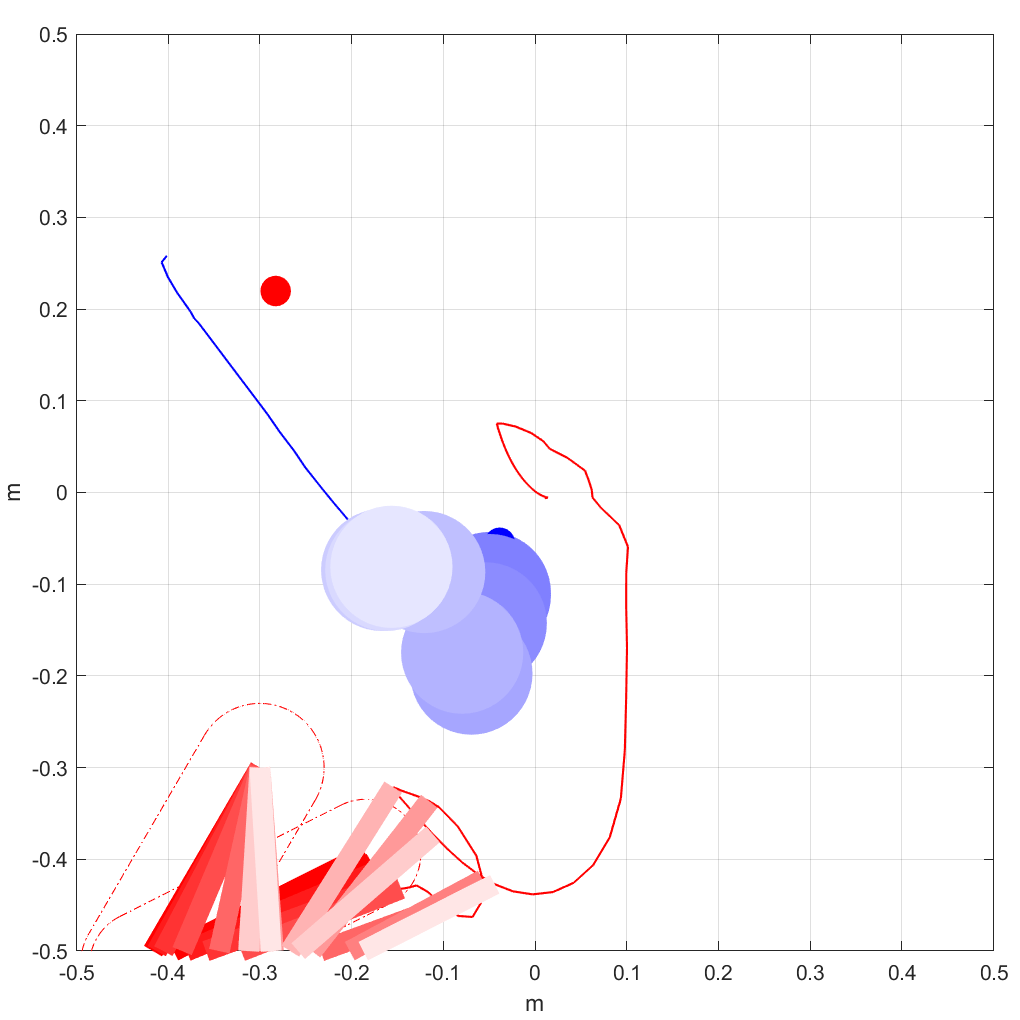}}
     \;
     \subfigure[\tiny M2,$\,$k$=200$\normalsize]{\label{fig:t2_ghat_phir_200graph}
       \includegraphics[width=0.725in]{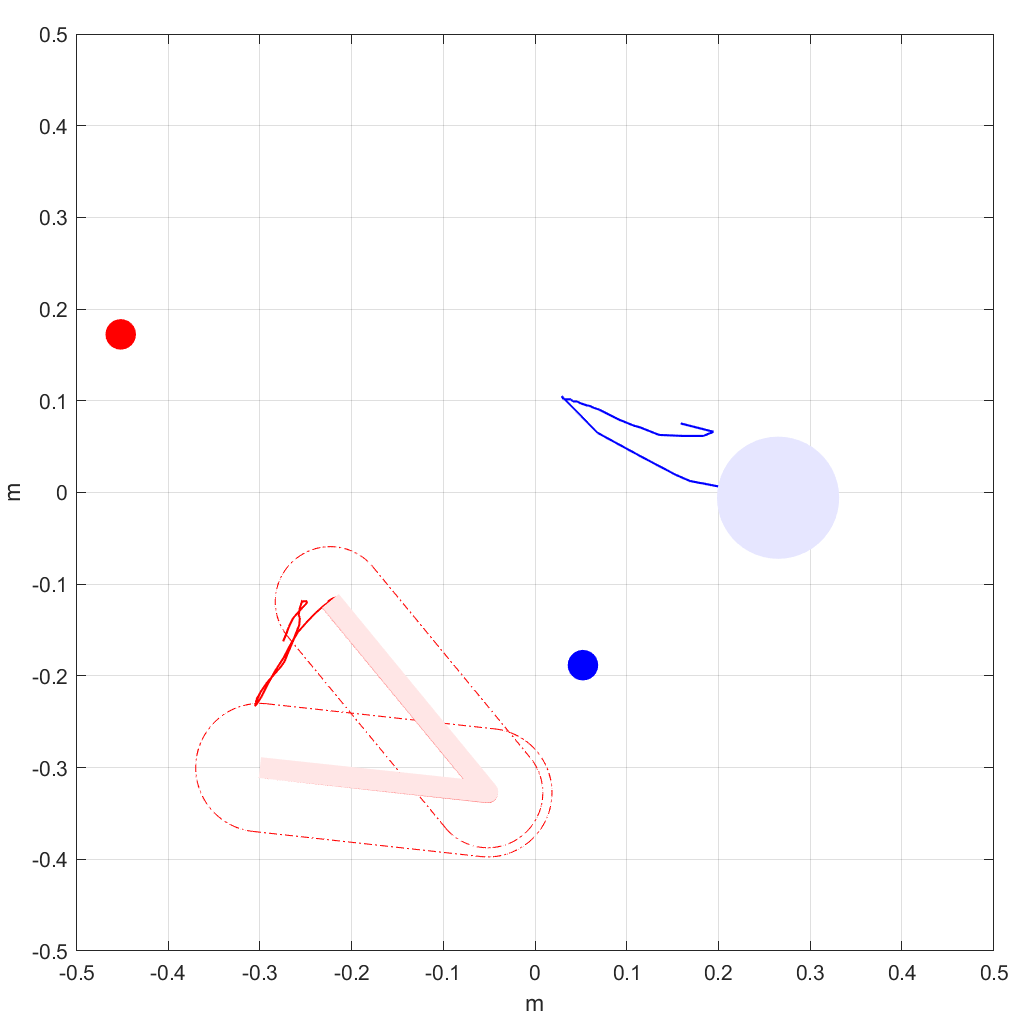}}   
     \;
     \subfigure[\tiny M2,$\,$k$=750$\normalsize]{\label{fig:t2_ghat_phir_750graph}%
         \includegraphics[width=0.725in]{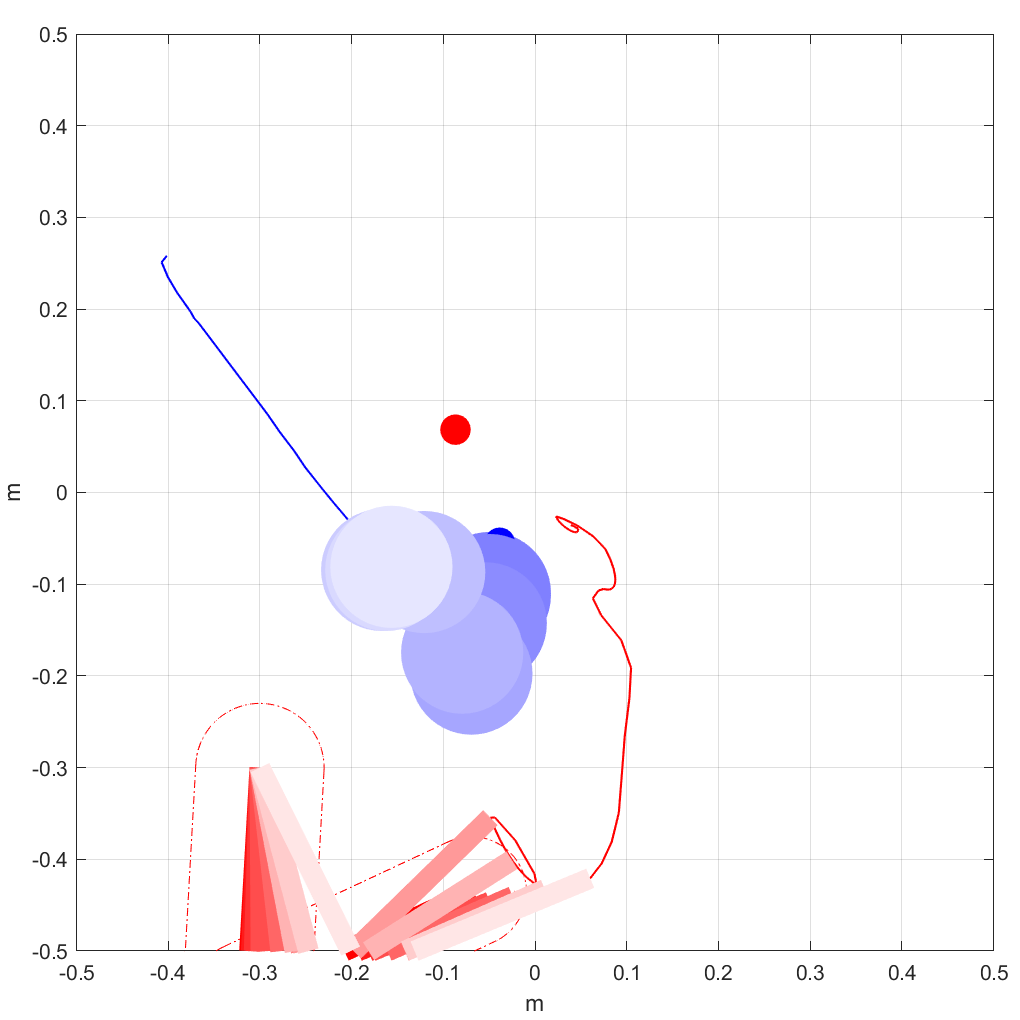}}
    \;
    \subfigure[\tiny M3,$\,$k$=200$\normalsize]{\label{fig:t2_gstr_phi_200graph}
        \includegraphics[width=0.725in]{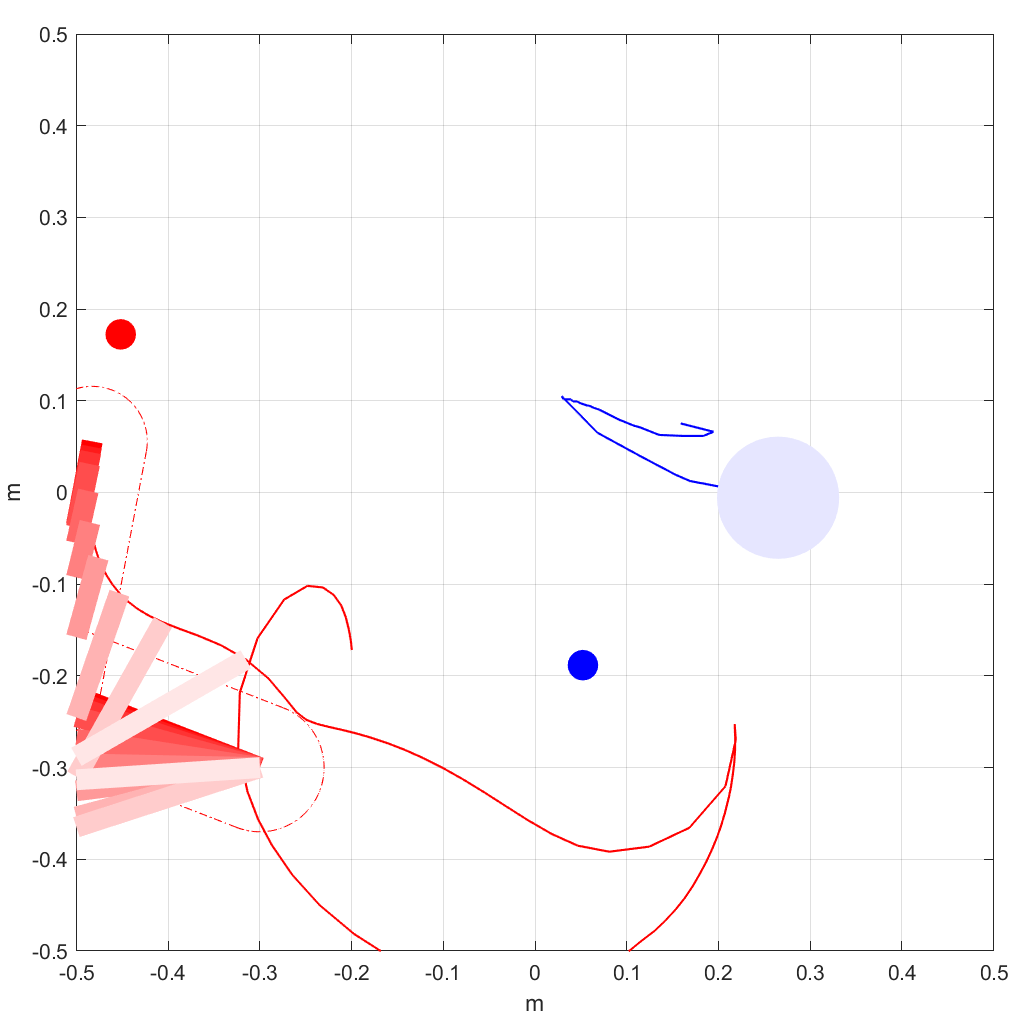}}
    \;
     \subfigure[\tiny M3,$\,$k$=750$\normalsize]{\label{fig:t2_gstr_phi_750graph}%
       \includegraphics[width=0.725in]{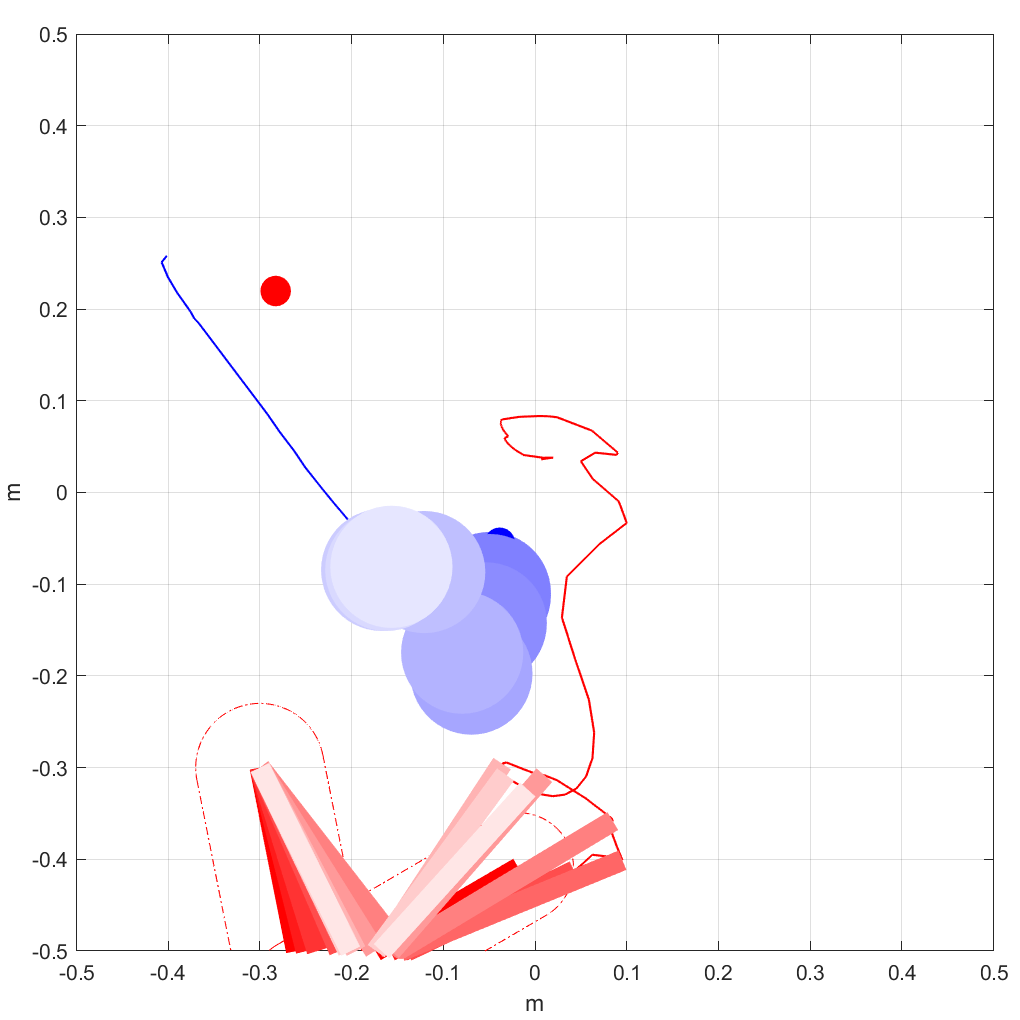}}   
     \;
     \subfigure[\tiny M4,$\,$k$=200$\normalsize]{\label{fig:t2_gstr_phir_200graph}%
     \includegraphics[width=0.725in]{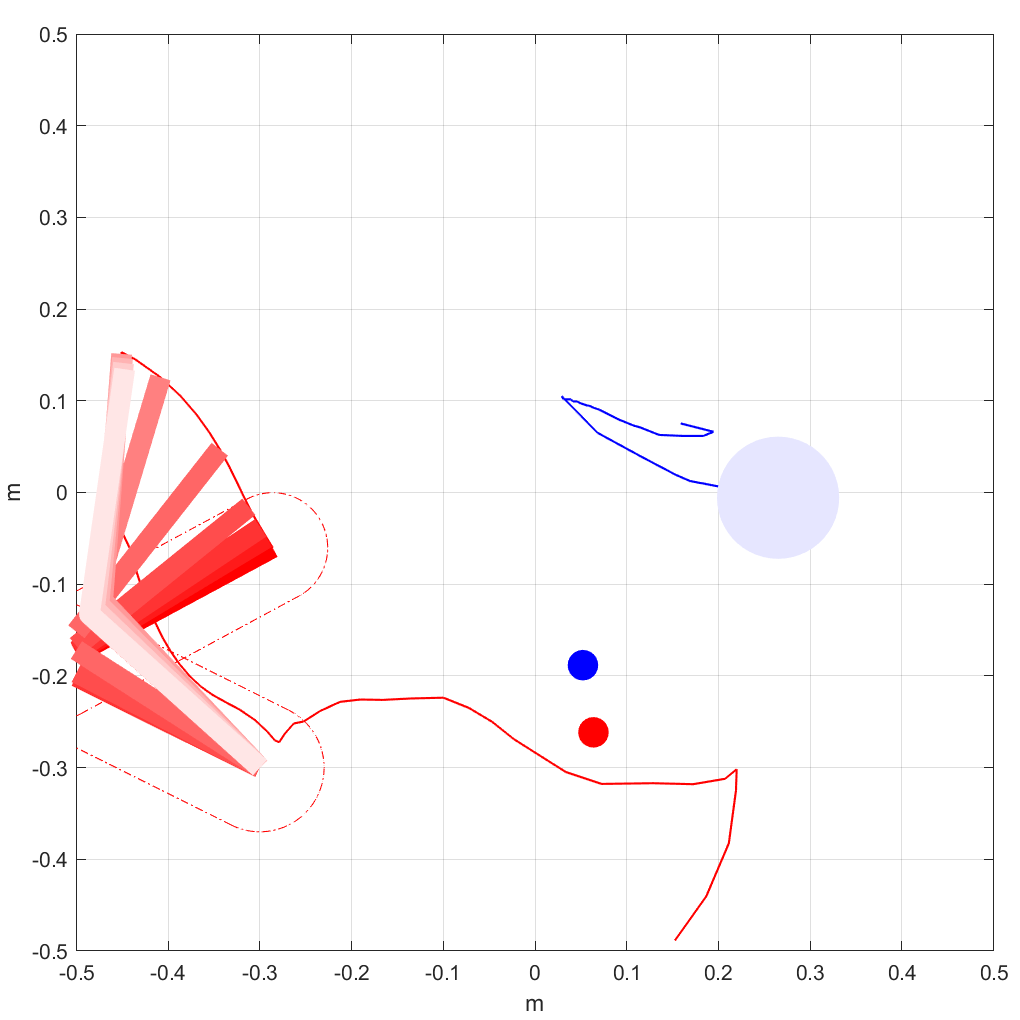}}
     \;
     \subfigure[\tiny M4,$\,$k$=750$\normalsize]{\label{fig:t2_gstr_phir_750graph}
       \includegraphics[width=0.725in]{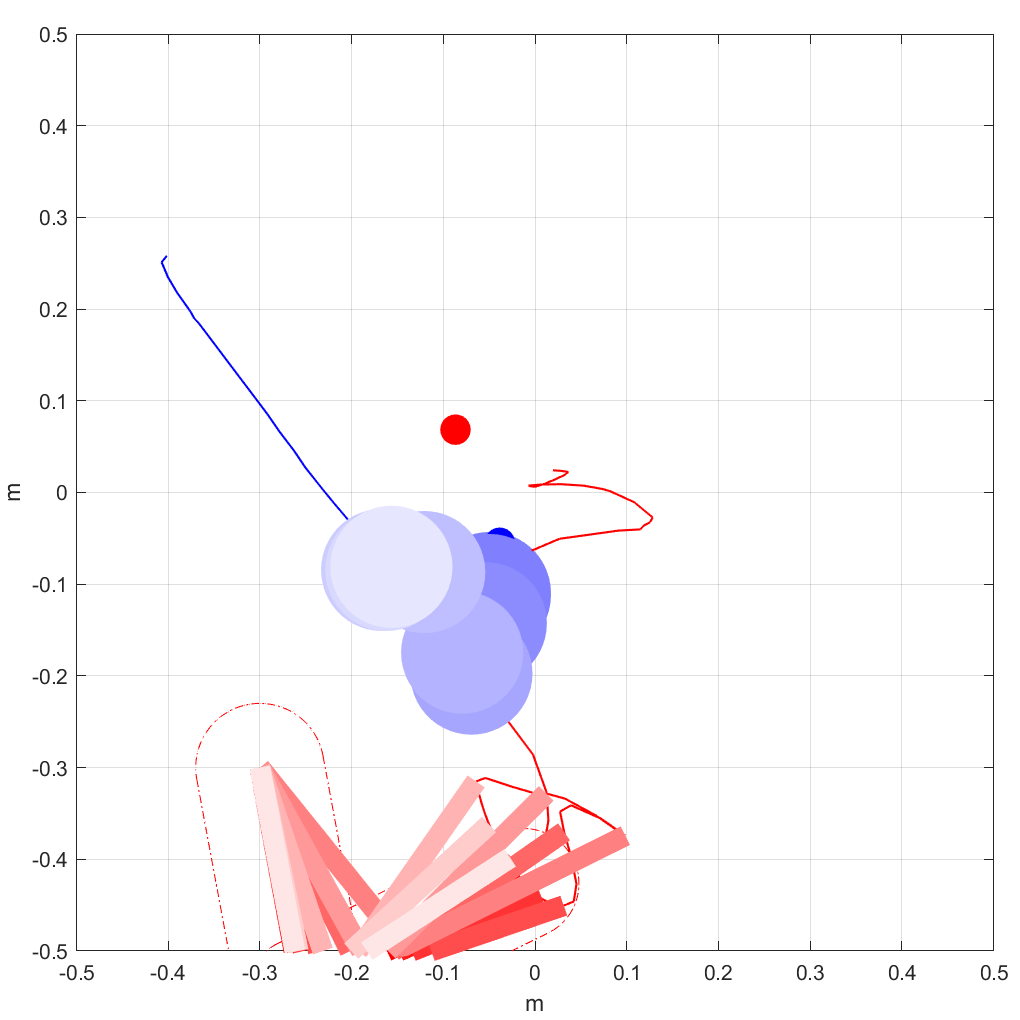}}
    }
\centering
  \caption{Example of simulated two-link robot manipulator trajectories from Trial 2. \label{traj2}}
\end{figure*}

We illustrate the performance of the RSSA in the following case study on a planar two-link robotic manipulator. Parametric uncertainty enters into the system model through the uncertainty in the dynamic parameters. This problem has been widely considered in literature. 
In \cite{SL_OACRM} it has been shown that by selecting appropriate payload and arm parameters, $\xi_1, \xi_2$, and $\xi_3$ below, the system dynamics may be made linearly dependent on the unknown parameters:
\begin{equation} \label{dyn1}
    \begin{split}
        &\begin{bmatrix} \xi_1 + 2\xi_2\cos(\theta_2) & \xi_3 + \xi_2\cos(\theta_2) \\   \xi_3 + \xi_2\cos(\theta_2) & \xi_3 \end{bmatrix}
        \begin{bmatrix} \Ddot{\theta}_1 \\ \Ddot{\theta}_2 \end{bmatrix}
        + \\
       &\begin{bmatrix} \xi_2(-2\sin(\theta_2)\dot{\theta}_1\dot{\theta}_2 - \sin(\theta_2)(\dot{\theta}_2)^2 \\  \xi_2\sin(\theta_2)(\dot{\theta}_1) \end{bmatrix}
       =
        \begin{bmatrix} \tau_1 \\ \tau_2 \end{bmatrix}       
    \end{split}_{\textstyle ,} 
\end{equation}
where $\theta_1, \text{ } \theta_2$ are the positions of the joints, $\dot{\theta}_1, \text{ } \dot{\theta}_2$ are the joint velocities, $\ddot{\theta}_1,  \text{ }\ddot{\theta}_2$ are the joint accelerations, and $\tau_1, \text{ } \tau_2$ are the torques applied at joint 1 and joint 2, respectively. We consider the control problem where the lengths of the different links are known, but the masses of the manipulator links are not. Here we assume the mass is evenly distributed along the length of each link, which results in a simplified formulation of the inertial terms. Finally, all unknown true values of the dynamic parameters lie within the known intervals. Using the upper and lower bounds on the masses and the known parameters, the upper and lower bounds for $\xi_1 \in [\underline{\xi_1}, \bar{\xi_1}], \xi_2 \in [\underline{\xi_2}, \bar{\xi_2}], \xi_3 \in [\underline{\xi_3}, \bar{\xi_3}]$ may be computed. These intervals define the family of functions, $\Sigma_g(x)$. 

\subsection{Design of the robust safety index} \label{subsection: DoRSI}
We use the same safety index 
\begin{equation} \label{eq: phistar}
    \phi = d_{min}^2 - d(t)^2 -k_1\dot{d}(t),
\end{equation}
 as in \cite{LT_CIASS}, where $d_{min}$ is the minimum distance requirement between the robot and obstacle, $d(t)$ is the real-time distance, $\dot{d}(t)$ is the relative velocity, and $k_1$ is a tunable constant. It has been shown in \cite{LT_CIASS} that this $\phi$ satisfies \Cref{def: sufficient}. This paper formulates $\phi_\alpha$ by considering the square of the difference between the expected and the estimated parameter value. The system safety index is defined as:
\begin{align} \label{sspec}
    \phi_{R} = d_{min}^2 - d(t)^2 -k_1\dot{d}(t) + k_{\xi}( \delta\xi^T \text{ } \Xi \text{ } \delta\xi),
\end{align}
 where $(\delta\xi^T \text{ } \Xi \text{ } \delta\xi)$ is used to evaluate parametric uncertainty. Specifically, $\delta\xi = {\xi^M_{i} - \hat{\xi}_i}$ is the difference between the center of the interval of $[\underline{\xi}, \bar{\xi}]$ and the estimate of $\xi$, $\Xi$ is a symmetric positive definite weighting matrix, and gain $k_{\xi}$ is a hyperparameter similar to $k_1$. The assumption behind this form is the expectation that the center point of the interval is approximate to the true value, but any interior value of the estimated parameter range could be used. The weighting term, $\Xi$, in \eqref{sspec} penalizes uncertainty in the parameters, generating a conservative system behavior for deviance from the expected value of the parameter. The following methodology for selecting the system safety index hyperparameters of \eqref{sspec} was leveraged for the case study.  First, a required minimum offset distance ($d_{min}$) is selected and then $k_1$ is tuned by observing the system response and penalizing the rate of change of the offset distance. Following this, the terms of \eqref{sspec} associated with $\phi_{\alpha}$ are normalized by the expected value of the parameter in ($\Xi$) and then are scaled by the amount of influence the designer wants the uncertainty to have on the system behavior via the magnitude of $k_{\xi}$. For the following simulations, $d_{min}=0.15$ meters ([m]), $k_1 = 0.01$ and $k_\xi = 20$. 
To apply \Cref{alg:RSSA}, we need to first verify that the conditions in \cref{thm:main} are satisfied. By design, $\phi_{R}$ meets conditions 3) and 4). We only need to verify conditions 1) and 2).
Consider the two-link manipulator control problem with the constraints discussed above. The prescribed problem meets 1) and we can use the adaptation method discussed in \cite{SL_OACRM} to estimate $\xi$, but 2) may not always be true for a class of dynamic systems. The proof of \Cref{lm2} is shown in \Cref{appdx:2}.


\begin{lemma}\label{lm2}[Positive Definiteness]
If 1) $\xi_1>2\xi_2$, 2) $\xi_2 > \xi_3$, and 3) $\Sigma_g(x)$ is sufficiently small, then \eqref{eq: find g*} is satisfied for manipulators with unknown link masses.
\end{lemma}

\subsection{Numerical Study}
To demonstrate the effectiveness of our proposed method in terms of generating safe controls under system parametric uncertainty, we conduct a set of numerical simulations where the manipulator completes goal reaching tasks while dodging obstacles. The simulation environment is shown in \Cref{fig:setup}, where the red two-link manipulator's objective is to reach the red goal (the small red circle), and the blue human-controlled cursor's objective is to reach the blue goal (the small blue circle). The grid provided as a backdrop to each image has a spacing of $0.1$ [m]. For each trial, the human and the robot are assigned multiple goals revealed sequentially upon reaching the previous goal, The human may observe the robot linkages and the robot may observe the cursor position to cooperatively reach the assigned goals. The simulation ends when the maximum number of simulation steps is reached (for presented trials, 1000 time steps). Time information is encoded in each image of the manipulator as the color gradient, from light-to-dark, reflecting the last ten time steps (e.g. the lightest manipulator and cursor position in \cref{fig:t2_ghat_phi_200graph} corresponds to time step $k=190$). The human is constantly moving, and we provide an estimation of human cursor position by adding a bounded noise to the ground truth human position. Even though the simulation is a discrete implementation of a continuous system, to align with the proposed theory our approach uses a continuous time formulation. However, a discrete-time safety index formulation as described in \cite{LT_CIASS} could also be used. The human's goal and the manipulator's goal are often nearby each other, which brings the manipulator in close proximity to the human-controlled cursor. The underlying adaptive controller uses the one defined in \cite{SL_OACRM}, as explained in \Cref{appendix: adaptive control law}. 
The underlying dynamic parameter values are given as: length of link 1: $l_1 = 0.25$ [m], length of link 2: $l_2 = 0.27$ [m], mass of link 1: $m_1 \in [26.75, 28.75]$ kilograms ([kg]), and the mass of link 2: $m_2 \in [13.30, 14.30]$ [kg]. 

To better understand the performance of RSSA and how the safety index in \eqref{sspec} affects that performance, four combinations of methods are compared across three separate trials. For a baseline safe control method, we leverage the safe control method presented in \cite{liu16}. This method can use any user-defined safety index so long as it meets the conditions in \Cref{def: sufficient}. Furthermore, the safe control value is synthesized as 
\begin{equation} \label{cliu16_cntl}
    u = \argmin_{u \in \mathcal{U}_\mathcal{S} } (u-u^r)^TQ(u-u^r),
\end{equation}
where $Q \in R^{m\times m}$ is positive definite. The four methods that are tested use different estimates of $g$ and different safety indices. Method 1 (M1) uses the safety index $\phi$ given in \eqref{eq: phistar} and $\hat{g}(x)$ from \eqref{slotine_aclaw} to compute $\mathcal{U}_\mathcal{S}$ in \eqref{cliu16_cntl}. Method 2 (M2) leverages the safety index $\phi_R$ given in \eqref{sspec} and $\hat{g}(x)$ from \eqref{slotine_aclaw} to compute $\mathcal{U}_\mathcal{S}$ in \eqref{cliu16_cntl}. Methods M3 and M4 use the proposed RSSA algorithm instead of \eqref{cliu16_cntl}. To obtain $\alpha^*$ in \eqref{rssa} and $g^*(x)$ in \eqref{eq: find g* equivalent problem}, the  minimax problem is solved via a discretization strategy in \cite{Rustem_Book} during the control loop. The calculated $\alpha^*$ and $L_{g^*}\phi$ are then used to calculate a safe control via \eqref{eq: final_law}. Method 3 (M3) uses a safety index $\phi$ given in \eqref{eq: phistar}, the parameter estimation algorithm from \eqref{slotine_aclaw} to adapt the parameters, and uses RSSA in \Cref{alg:RSSA} with $\phi_\alpha=0$ to synthesize the safe control. Finally, Method 4 (M4) uses a safety index given in \eqref{sspec}, the parameter estimation algorithm given in \eqref{slotine_aclaw}, and RSSA for synthesizing the safe control. The three trails use three pre-recorded human trajectories. An example of the simulated trajectories of the four compared methods at two simulation time steps ($k=200$, $k=750$) are shown in \cref{traj2}. In practice, we clip the control output to compensate the discrepancy between discrete-time system and continuous-time system theory, as the same control force will result different movements in two systems. Note that the theory is based on continuous-time systems, whereas the simulation is a discrete-time system. The control limit is placed at $\pm20$ [N$\cdot$m].


\begin{table*}
\label{table-of-results}
\caption{Simulation Results for a Two-link manipulator, number of Goals reached (GOAL), safety distance violations (VIOL), and minimum safety distance (DIST).} 
\begin{tabularx}{\textwidth}{@{}l*{20}{C}c@{}}
\toprule
\textbf{}& \multicolumn{3}{|>{\hsize=\dimexpr3\hsize+3\tabcolsep+\arrayrulewidth\relax}C}{Trial 1}& \multicolumn{3}{|>{\hsize=\dimexpr3\hsize+3\tabcolsep+\arrayrulewidth\relax}C}{Trial 2}&\multicolumn{3}{|>{\hsize=\dimexpr3\hsize+3\tabcolsep+\arrayrulewidth\relax}C}{Trial 3}\\
\midrule
METHODS     & GOAL & VIOL & DIST & GOAL & VIOL & DIST & GOAL & VIOL & DIST \\ 
\midrule
NO OBSTACLE                     & 17      & 0          & 0.6160    & 17    & 0    & 0.6160   & 17     & 0     & 0.6160   \\
M0: $\hat{g}_{BAD}$(x), $\phi(x)$ & 1       & 0          & 0.1893    & 7     & 0    & 0.2194   & 4      & 7     & 0.1196   \\
M1: $\hat{g}$(x), $\phi(x)$       & 1       & 0          & 0.1864    & 7     & 0    & 0.2265   & 4      & 0     & 0.2069    \\ 
M2: $\hat{g}$(x), $\phi_R(x)$     & 3       & 0          & 0.1841    & 5     & 0    & 0.2399   & 3      & 0     & 0.2173    \\ 
M3: $g^*(x)$, $\phi(x)$           & 3       & 0          & 0.1501    & 7     & 0    & 0.2303   & 4      & 0     & 0.1542    \\ 
M4: $g^*(x)$, $\phi_R(x)$         & 1       & 0          & 0.1849    & 4     & 0    & 0.1788   & 1      & 0     & 0.2001    \\ 
\bottomrule
\end{tabularx}
\end{table*}

To measure the safety and efficiency of the system, we consider an approach following the spirit of the work in \cite{TC_SCAUEF}, which defines a joint metric for the efficiency and safety of a system. For the efficiency score, we consider the number of goals reached within fixed time steps ($k=1000$). For the safety score, we consider the smallest distance between the manipulator and the environmental obstacle and the number of violations that occur. Here, a violation is defined when an obstacle is closer to the robot than the defined the minimum safe distance. We propose two main hypotheses, which will be discussed in the following subsections:

\begin{hypothesis}\label{H1}
The safety control performance of methods that directly leverage estimates of the dynamic parameters is dependent on the $g(x)$ parameter estimation error.
\end{hypothesis}
\begin{hypothesis}\label{H2}
Safety index $\phi_R$ increases the average safety distance between the robot and the closest obstacle.
\end{hypothesis}

\subsection{Results and Discussion}
A comparison of the results of the four methods across different trials and the results from a no-obstacle run of the controller (which indicates the performance boundary) is summarized in Table \ref{table-of-results}. 

\subsubsection{Discussion of \cref{H1}}
To demonstrate the effects of a bad estimation of $g(x)$, method M0 is an exact copy of method M1, except that the estimation algorithm holds the parameters at a random constant value over the course of the experiment. While Trial(s) 1 and 2 seemingly suggests that naively leveraging the estimates of the system parameters can be used for the safe control of the manipulator, the violations that occurs in Trial 3 are due to this bad estimation as the performance of methodology M1 does not suffer from the same violations. However, if a theoretical estimation algorithm was devised which could recover the exact dynamic parameters of the system at all instances in time, this would be equivalent to knowing the system model. Therefore, the performance of the system on maintaining a safe system is dependent on the quality of the estimates of the estimation algorithm, which is reflected by these violations appearing in Trial 3.

\subsubsection{Discussion of \cref{H2}}
Simply comparing how these systems behave (such as just looking at the minimum safe distance obtained over the run) does not provide sufficient characterization of the performance of the system. As such, the minimum safe distance from each system composition is provided over the length of each Trial in \cref{min_safe_dist}. In Trials 1 and 3, $\phi_R$ increases the minimum safe distance across all runs, but this is not reflected in Trial 2. However, the average safety distance over the course of the trial increases on a per method basis, as seen in Table \ref{table-dps}. 


To qualitatively describe the tests, at the start time, the human cursor is initialized in the top right corner of the frame, making the human beyond the minimum safety distance and thus not triggering the safety controller. However, as the human and manipulator come closer together, as in \cref{fig:t2_ghat_phi_750graph}, the manipulator changes its control as to not strike the human. Oftentimes, this safety behavior generalizes to causing the manipulator to remain stationary, as in what occurs in \cref{fig:t2_ghat_phir_200graph}, or causing the manipulator to travel in the direction opposite of the human, as shown in \cref{fig:t2_gstr_phir_750graph} until either the human or the robot retreats to a safe distance. In Trial 2 time step $k=750$, shown in \cref{fig:t2_ghat_phi_750graph}, \cref{fig:t2_gstr_phi_750graph}, \cref{fig:t2_ghat_phir_750graph}, \cref{fig:t2_gstr_phir_750graph}, the human makes a sharp move towards the manipulator which causes it to abruptly change its position and move away from the human.

\begin{table}[hb]
\begin{center}
\caption{Average Safety Distance over Trials}\label{table-dps}
\vspace{-10pt}
\begin{tabular}{cccc}
\hline
METHOD & TRIAL 1 [m] & TRIAL 2 [m] & TRIAL 3 [m] \\\hline
M1: $\hat{g}$(x), $\phi(x)$     & 0.3921       & 0.5081          & 0.4141   \\ 
M2: $\hat{g}$(x), $\phi_R(x)$   & 0.4318       & 0.5222          & 0.4449   \\ 
M3: $g^*(x)$, $\phi(x)$         & 0.4007       & 0.4977          & 0.4064    \\ 
M4: $g^*(x)$, $\phi_R(x)$       & 0.4171       & 0.5175          & 0.4394    \\  \hline
\end{tabular}
\end{center}
\end{table}


With acceptable estimation accuracy, no safety violations occur in any of the trials. The advantage of using RSSA methods (M3 and M4) is that one gains system safety guarantees, which are not guaranteed with the naive methods (M1 and M2). The use of safety index $\phi_R$ has the general effect of dropping efficiency performance, but has the advantage of increasing average safety distance.

\section{CONCLUSIONS AND LIMITATIONS } \label{sec: conc}
In this paper, we present a method to design a safe controller for systems with parametric uncertainty in the system model called RSSA. A numerical study with a planar two-link robot is conducted and the simulation results indicate the effectiveness of the proposed method in terms of providing provably safe control under system parametric uncertainty. The proposed method implementation is publicly available at \url{https://github.com/intelligent-control-lab/ARSSA}.

Future work will look at generalizing the work presented in this document to additional dynamical systems. We will further conduct a sensitivity analysis of the influence of the parameters of the safety index on the RSSA method. Finally, the application of our safe controller on hardware shall require the observation of additional hardware limits, such as joint actuator limits. A safe controller that respects actuator limits may be derived \cite{wzhao_anonymous2021modelfree} or directly learned \cite{twei_anonymous2021safe} if there are control constraints. The paper with quoted appendices may be found at \url{https://arxiv.org/abs/1912.09095}.


\bibliography{ifacconf} 

\begin{thebibliography}{4}
\providecommand{\natexlab}[1]{#1}
\providecommand{\url}[1]{\texttt{#1}}
\providecommand{\urlprefix}{URL }
\expandafter\ifx\csname urlstyle\endcsname\relax
  \providecommand{\doi}[1]{doi:\discretionary{}{}{}#1}\else
  \providecommand{\doi}{doi:\discretionary{}{}{}\begingroup
  \urlstyle{rm}\Url}\fi

\bibitem[{Able(1956)}]{Abl:56}
Able, B. (1956).
\newblock Nucleic acid content of microscope.
\newblock \emph{Nature}, 135, 7--9.

\bibitem[{Able et~al.(1954)Able, Tagg, and Rush}]{AbTaRu:54}
Able, B., Tagg, R., and Rush, M. (1954).
\newblock Enzyme-catalyzed cellular transanimations.
\newblock In A.~Round (ed.), \emph{Advances in Enzymology}, volume~2, 125--247.
  Academic Press, New York, 3rd edition.

\bibitem[{Keohane(1958)}]{Keo:58}
Keohane, R. (1958).
\newblock \emph{Power and Interdependence: World Politics in Transitions}.
\newblock Little, Brown \& Co., Boston.

\bibitem[{Powers(1985)}]{Pow:85}
Powers, T. (1985).
\newblock Is there a way out?
\newblock \emph{Harpers}, 35--47.

\end{thebibliography}


\begin{thebibliography}{14}
\providecommand{\natexlab}[1]{#1}
\providecommand{\url}[1]{\texttt{#1}}
\providecommand{\urlprefix}{URL }
\expandafter\ifx\csname urlstyle\endcsname\relax
  \providecommand{\doi}[1]{doi:\discretionary{}{}{}#1}\else
  \providecommand{\doi}{doi:\discretionary{}{}{}\begingroup
  \urlstyle{rm}\Url}\fi

\bibitem[{Ames. et~al.(2019)Ames., Coogan, Egerstedt, Notomista, Sreenath, and
  Tabuada}]{Ames_2019}
Ames., A., Coogan, S., Egerstedt, M., Notomista, G., Sreenath, K., and Tabuada,
  P. (2019).
\newblock Control barrier functions: Theory and applications.
\newblock \emph{2019 18th European Control Conference (ECC)}.

\bibitem[{Anon1(2021{\natexlab{a}})}]{wzhao_anonymous2021modelfree}
Anon1 (2021{\natexlab{a}}).
\newblock Model-free safe control for zero-violation reinforcement learning.
\newblock In \emph{Submitted to 5th Annual Conference on Robot Learning}.
\newblock \urlprefix\url{https://openreview.net/forum?id=UGp6FDaxB0f}.
\newblock Under review.

\bibitem[{Anon1(2021{\natexlab{b}})}]{twei_anonymous2021safe}
Anon1 (2021{\natexlab{b}}).
\newblock Safe control with neural network dynamic models.
\newblock In \emph{Submitted to 5th Annual Conference on Robot Learning}.
\newblock \urlprefix\url{https://openreview.net/forum?id=TBor1Fsz812}.
\newblock Under review.

\bibitem[{Isidori(2013)}]{isidori2013nonlinear}
Isidori, A. (2013).
\newblock \emph{Nonlinear control systems}.
\newblock Springer Science \& Business Media.

\bibitem[{Liu and Tomizuka(2014)}]{LT_CIASS}
Liu, C. and Tomizuka, M. (2014).
\newblock Control in a safe set: Addressing safety in human-robot interactions.
\newblock In \emph{Proceedings of the ASME 2014 Dynamic Systems and Control
  Conference}. ASME, San Antonio, TX, USA.

\bibitem[{Liu and Tomizuka(2015)}]{L_SEA}
Liu, C. and Tomizuka, M. (2015).
\newblock Safe exploration: Addressing various uncertainty levels in human
  robot interactions.
\newblock In \emph{2015 American Control Conference (ACC)}. IEEE, Chicago, IL,
  USA.

\bibitem[{{Liu} and {Tomizuka}(2016)}]{liu16}
{Liu}, C. and {Tomizuka}, M. (2016).
\newblock Algorithmic safety measures for intelligent industrial co-robots.
\newblock In \emph{2016 IEEE International Conference on Robotics and
  Automation (ICRA)}, 3095--3102.

\bibitem[{Rauch and Smoller(1978)}]{rauch1978qualitative}
Rauch, J. and Smoller, J.A. (1978).
\newblock Qualitative theory of the fitzhugh-nagumo equations.

\bibitem[{Rustem and Howe(2003)}]{Rustem_Book}
Rustem, B. and Howe, M. (2003).
\newblock \emph{Algorithms for Worst-Case Design and Applications to Risk
  Management}.
\newblock Princeton University Press, 41 William Street, Princeton, New Jersey
  08540, United States.

\bibitem[{Slotine and Li(1987)}]{SL_OACRM}
Slotine, J. and Li, W. (1987).
\newblock On the adaptive control of robot manipulators.
\newblock In \emph{The International Journal of Robotics Research}, volume~6,
  49--59. IEEE.

\bibitem[{Slotine and Li(1991)}]{JJS_book}
Slotine, J. and Li, W. (1991).
\newblock \emph{Applied Nonlinear Control}.
\newblock Prentice Hall.

\bibitem[{{Taylor} and {Ames}(2020)}]{TA_ASCBF}
{Taylor}, A.J. and {Ames}, A.D. (2020).
\newblock Adaptive safety with control barrier functions.
\newblock In \emph{2020 American Control Conference (ACC)}, 1399--1405.

\bibitem[{Taylor et~al.(2020)Taylor, Singletary, Yue, and
  Ames}]{pmlr-v120-taylor20a}
Taylor, A., Singletary, A., Yue, Y., and Ames, A. (2020).
\newblock Learning for safety-critical control with control barrier functions.
\newblock In \emph{Proceedings of the 2nd Conference on Learning for Dynamics
  and Control}, Proceedings of Machine Learning Research. PMLR.

\bibitem[{Wei and Liu(2019)}]{TC_SCAUEF}
Wei, T. and Liu, C. (2019).
\newblock Safe control algorithms using energy functions: A unified framework,
  benchmark, and new directions.
\newblock In \emph{58th Conference on Decision and Control}. IEEE.

\end{thebibliography}

\newpage

\appendix

\begin{figure*}
{
    \subfigure[Trial 1: Distance to closest point]{\label{fig:t1_safedist}%
        \includegraphics[width=2in, height=1.50in]{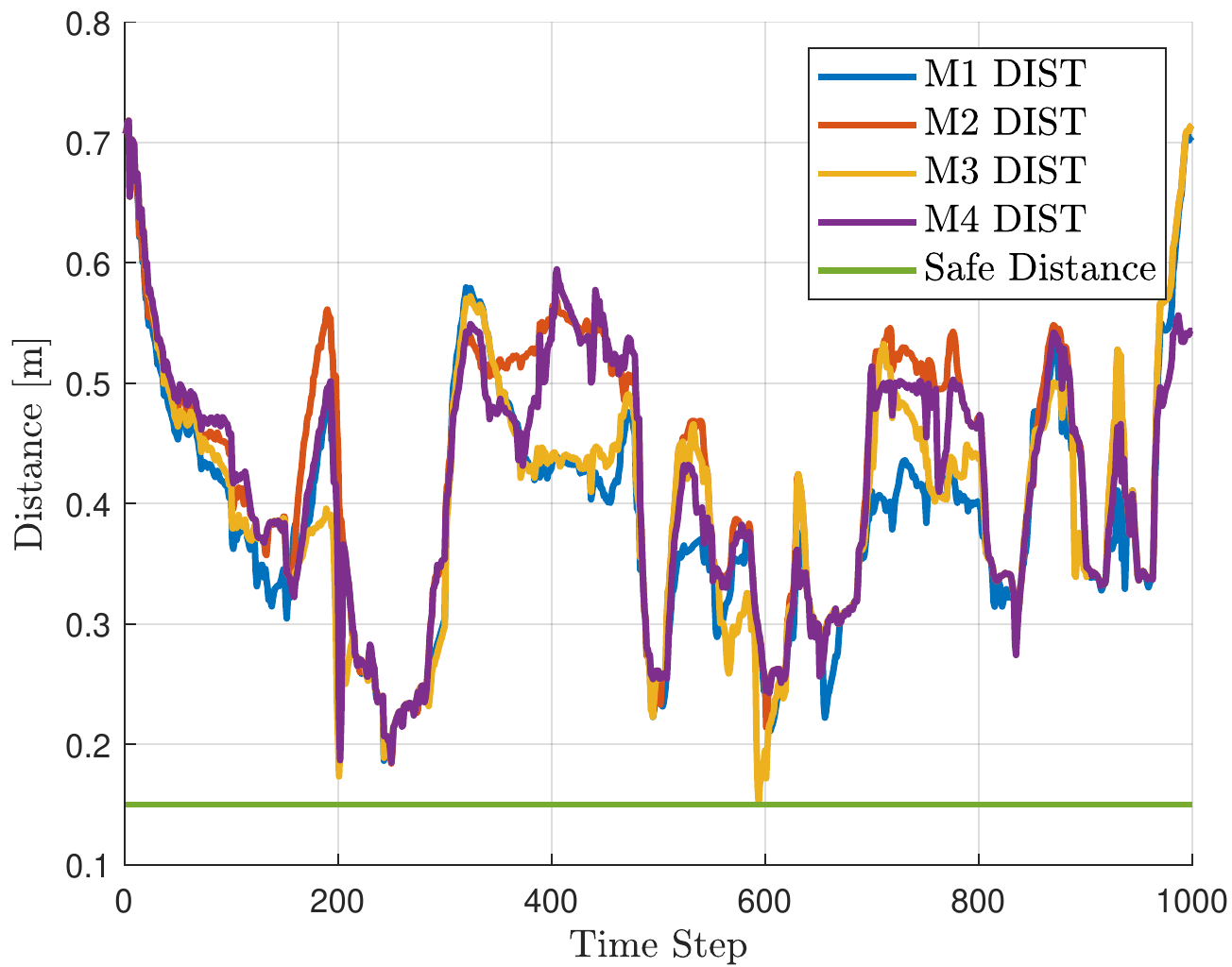}}
    \;
    \subfigure[Trial 2: Distance to closest point]{\label{fig:t2_safedist}%
        \includegraphics[width=2in, height=1.50in]{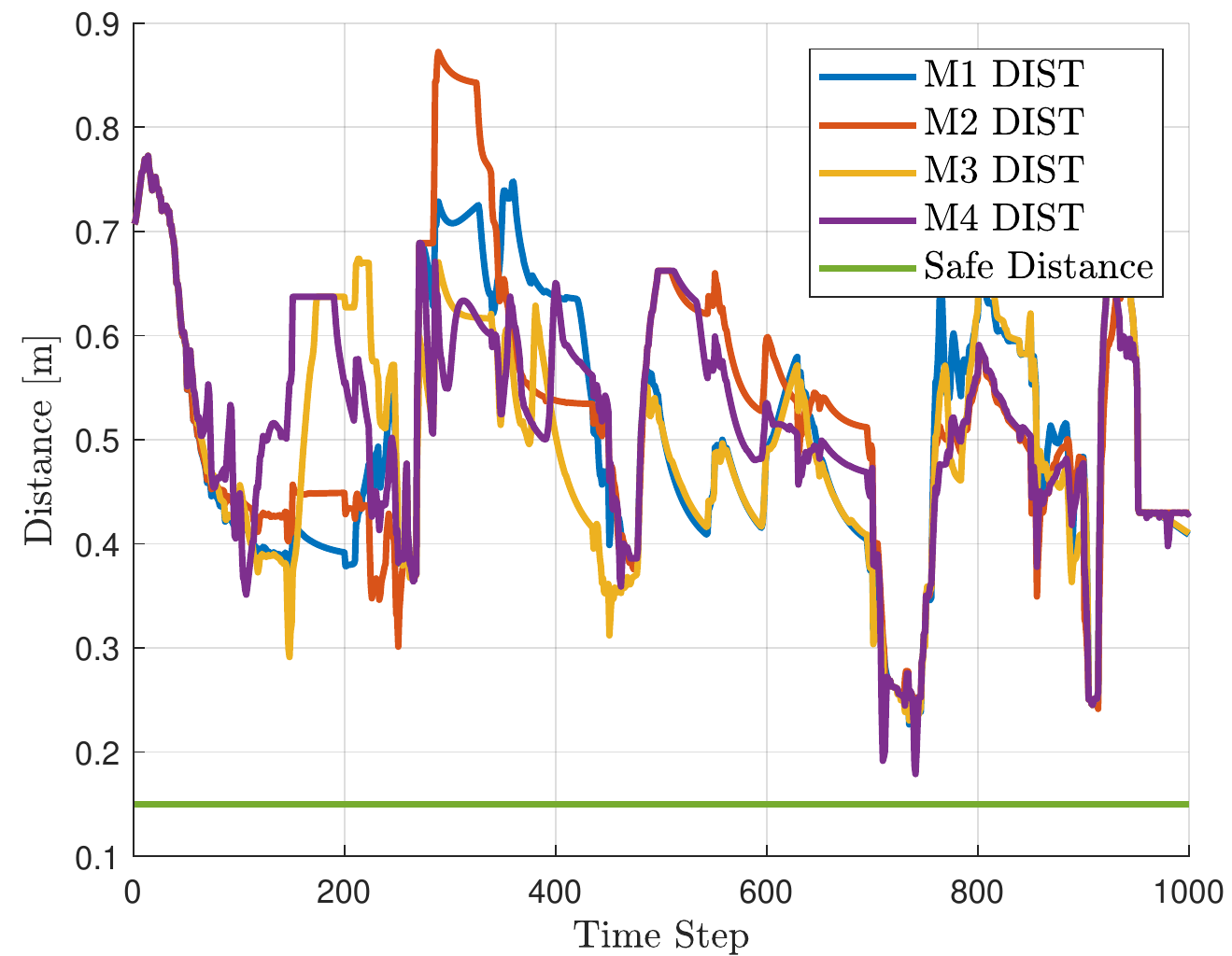}}
    \;
    \subfigure[Trial 3: Distance to closest point]{\label{fig:t3_safedist}%
        \includegraphics[width=2in, height=1.50in]{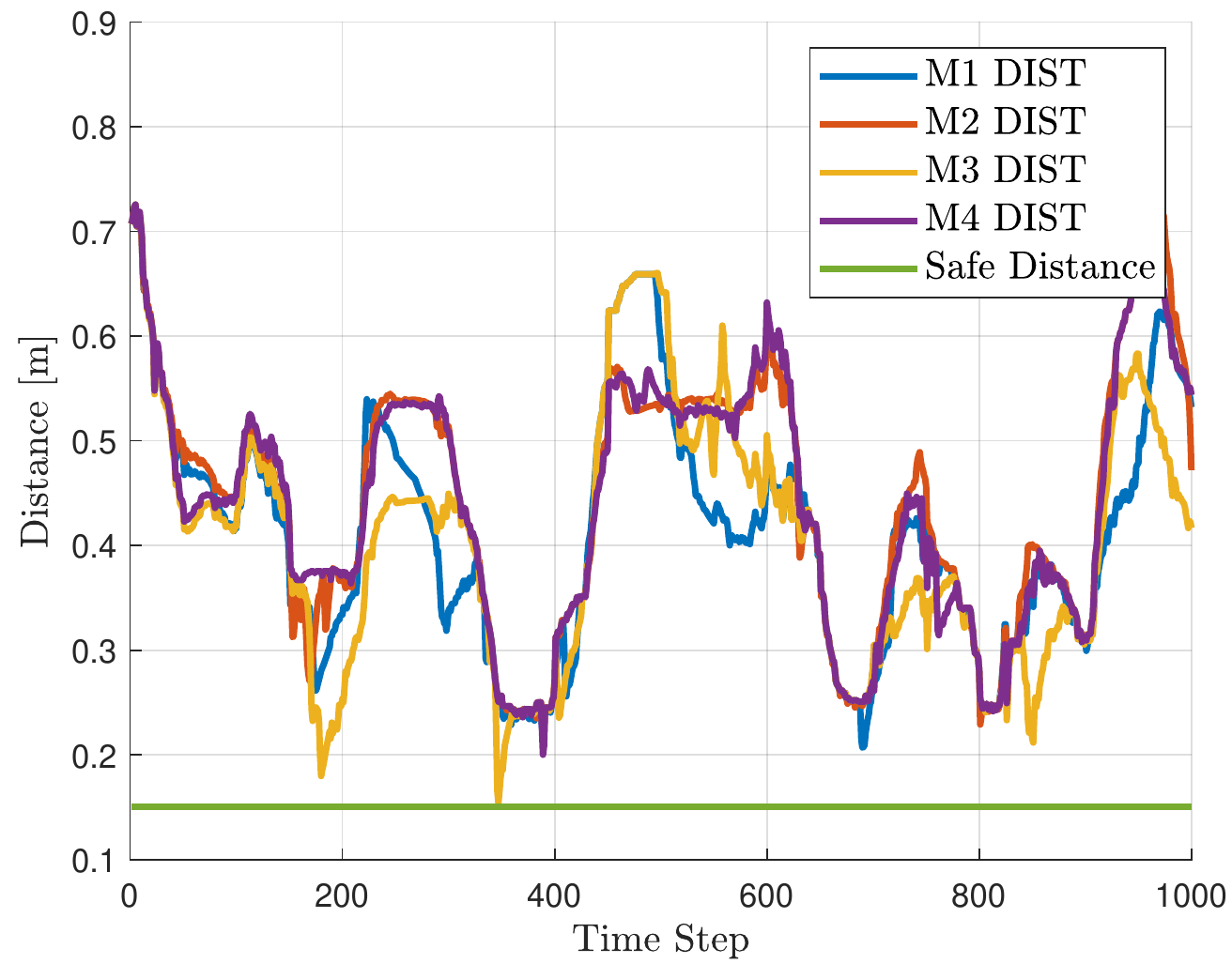}}        
}
\centering
  \caption{Simulated two-link robot manipulator, manipulator joint velocities, and distance to closest point.\label{min_safe_dist}}
\end{figure*}

\section{Proof of Lemma 1} \label{PCSID}
\begin{pf}
Note that $X(\phi + \phi_{\alpha})\subseteq \mathcal{L}(\phi + \phi_{\alpha})$. Suppose that $X(\phi + \phi_{\alpha}) \not\subset \mathcal{X}_S$, then there exists a time $t_1$ that the system is on the boundary of the safe set and will enter into the unsafe set. This corresponds to the conditions: 
\begin{equation}\label{eq: condition in pf}
  \phi(x(t_1)) + \phi_{\alpha}(t_1)\leq 0, \phi_0(x(t_1))=0, \dot\phi_0(x(t_1))>0. 
\end{equation}
Since $\phi_\alpha$ is positive semi-definite, \eqref{eq: condition in pf} implies that $\phi(x(t_1))\leq 0$. 
The condition $X(\phi)\subset\mathcal{X}_S$ implies that when $\phi_0(x(t_1))=0$ and $\phi(x(t_1))\leq 0$, the system trajectory under the condition \eqref{eq: phi dot} should always ensure that $\dot\phi_0(x(t_1))\leq 0$. Since the condition \eqref{eq: phi dot} only concerns the case that $\phi\geq 0$, then to ensure $X(\phi)\subset\mathcal{X}_S$, either of the following conditions should hold:
\begin{equation}\label{eq: condition 1}
    \begin{cases}
    \phi_0(x(t_1))=0 &\multirow{2}{*}{$\mathlarger{ \mathlarger{\mathlarger{\boldsymbol\implies}}} \dot\phi_0(x(t_1))\leq 0$}\\
    \phi(x(t_1))<0,
    \end{cases}
\end{equation}
\begin{equation}\label{eq: condition 2}
    \begin{cases}
    \phi_0(x(t_1))=0 &\multirow{3}{*}{$ \mathlarger{ \mathlarger{\mathlarger{\boldsymbol\implies}}}\dot\phi_0(x(t_1))\leq 0$}\\
    \phi(x(t_1))=0 \\
    \dot\phi(x(t_1))\leq -\eta.
    \end{cases}
\end{equation}
When $\phi(x(t_1))<0$, according to \eqref{eq: condition 1}, $\dot\phi_0(x(t_1))\leq 0$, which contradicts with \eqref{eq: condition in pf}.

When $\phi(x(t_1)) = 0$, then the positive semi-definiteness of $\phi_\alpha$ and $\phi(x(t_1)) + \phi_{\alpha}(t_1)\leq 0$ imply $\phi_\alpha(t_1) = \phi(x(t_1)) + \phi_{\alpha}(t_1) = 0$. The condition in \eqref{eq: phi dot} is triggered for the combined safety index $\phi+\phi_\alpha$ where $\dot\phi(x(t_1)) + \dot\phi_\alpha(t_1) \leq -\eta$. Since $\phi_\alpha$ is a smooth positive function on time, $\phi_\alpha(t_1)=0$ implies that $\dot\phi_\alpha(t_1)\geq 0$. Hence $\dot\phi(x(t_1)) \leq -\eta$. According to \eqref{eq: condition 2}, $\dot\phi_0(x(t_1))\leq 0$, which contradicts with \eqref{eq: condition in pf}. Then:
\begin{align}
&\exists \text{ } t_1 \text{ } | \text{ } \phi(x(t_1)) + \phi_{\alpha}(t_1) \leq 0, \phi_0(x(t_1)) = 0, \nonumber \\
&\text{ and } \dot{\phi}_0(x(t_1)) > 0 
\end{align}
However, as:
\begin{align}
    X(\phi) \subset \mathcal{X}_S
\end{align}
Then the controller on $\phi$ should take action such that whenever:
\begin{align*}
    \phi_0(x(t_1)) = 0 \text{ and } \phi(x(t_1))\leq0 \implies \dot{\phi}_0(x(t_1)) < 0
\end{align*}
Therefore, in the presence of a controller on $\phi$, only two conditions can hold:
\begin{enumerate}
    \item $\phi_0(x(t_1)) = 0$ and $\phi(x(t_1))< 0 \implies \dot{\phi}_0(x(t_1)) < 0$
    \item $\phi_0(x(t_1)) = 0$, $\phi(x(t_1)) = 0$ and $\dot{\phi}(x(t_1)) < -\eta \implies \dot{\phi}_0(x(t_1)) < 0$
\end{enumerate}
\begin{align*}
\intertext{So if $\phi_{\alpha}(t_1) > 0$:}
    \phi_{\alpha}(t_1) > 0 \text{ and } \phi(x(t_1)) + \phi_{\alpha}(t_1) \leq 0 \implies \phi(x(t_1)) \leq 0
\end{align*}
\begin{align*}
\intertext{Or if $\phi_{\alpha}(t_1) = 0$, the derivative constraint is triggered when:}
    \dot{\phi}(x(t_1)) + \dot{\phi}_{\alpha}(t_1) \leq -\eta
\end{align*}
\begin{align*}
\intertext{However, as $\phi_{\alpha}$ is a smooth positive function of time,}
    \phi_{\alpha}(t_1) = 0 \implies \dot{\phi}_{\alpha}(t_1) \geq 0\\
    \implies \dot{\phi}(x(t_1)) \leq -\eta
\end{align*}
Therefore both of these cases violate the above assumption and the statement is true.
\end{pf}
\section{Proof of Lemma 2}\label{appendix: lemma 2 proof}
\begin{pf}
Pick $g^*(x)\in\Sigma_g(x)$ and define $u = - c L_{g^*}\phi$ where $c$ is a constant to be picked. We are going to show that there must exist a $c$ that meets the inequality in \eqref{eq: robust safe set in g}. Plugging $u$ to the inequality, we have
\begin{align}
    L_f\phi + L_g\phi\cdot u &= L_f\phi - c L_g\phi\cdot L_{g^*}\phi\\
    &\leq L_f\phi - c \alpha \|L_g\phi\|\|L_{g^*}\phi\| \nonumber\\
    &\leq L_f\phi - c \alpha\beta \|L_{g^*}\phi\| \nonumber.
\end{align}
The two inequalities follow from the conditions in \eqref{eq: uncertainty requirement for g}. Hence as long as we choose
\begin{align}\label{eq: condition c}
    c\geq \frac{L_f\phi +\eta(t)}{\alpha\beta \|L_{g^*}\phi\|},
\end{align}
the inequality in \eqref{eq: robust safe set in g} holds and hence \eqref{eq: robust safe set in g} is non-empty.
\end{pf}

\section{Proof of Lemma 3}\label{appx3}
\begin{pf}
When $L_f\phi+\eta(t) \leq 0$, it is straightforward to see that the optimal solution of \eqref{eq: find g*} is $u=0$. In the following discussion, we derive for the case that $L_f\phi+\eta(t) >0$. Let $u = -c \frac{L_{g^*}\phi}{\|L_{g^*}\phi\|}$, where both $c$ and $g^*(x)$ must be optimized. Note that $\|u\| =c$. For a given $g^*(x)$, according to the constraints in \eqref{eq: find g*}, the smallest possible $c$ must satisfy:
\begin{align}\label{eq: finding c}
    c \cdot \min_{g(x)\in\Sigma_g(x)} L_g\phi\cdot \frac{L_{g^*}\phi}{\|L_{g^*}\phi\|} = {L_f\phi +\eta(t)}.
\end{align}
To minimize $c$, we pick $g^*(x)$ to maximize the multiplier to $c$ on the left hand side of \eqref{eq: finding c}. Hence we should pick $g^*(x)$ that solves \eqref{eq: find g* equivalent problem}. For the chosen $g^*(x)$, define $\alpha^* := \min_{g(x)\in\Sigma_g(x)} {L_{g^*}\phi \cdot L_g\phi}$. Hence the minimum norm of $u$ is
\begin{align} \label{eqn: min_control_effort}
    \|u\| = c = {\frac{L_f\phi +\eta(t)}{\alpha^*}} \|L_{g^*}\phi\|.
\end{align}
And then $u$ satisfies \eqref{rssa}. The claim in \cref{lemma: minimum effort} is verified.
\end{pf}

\section{Proof of Lemma 4}
Consider the Lie derivative of the above system with respect to a properly designed safety index:
\begin{equation} \label{lie_derivative_2link}
\begin{split}
    L_g\phi = \frac{[\Delta_x, \Delta_y]}{d(t)}J_{M}M^{-1} = \begin{bmatrix} a_1 & a_2  \end{bmatrix} M^{-1}
    \end{split}, 
\end{equation}
where the vector $[\Delta_x, \Delta_y]$ represents the relative position between the closest point on the obstacle and the manipulator, $J_m$ is the Jacobian at the closest point on the manipulator, and $M^{-1}$ is the inverse of the manipulator inertia matrix. 
To verify \eqref{eq: uncertainty requirement for g}, we just need to check the following equality:
\begin{equation} \label{sec3:simplified_pd_cond}
\begin{split}
    L_{g_1}\phi \cdot L_{g_2}\phi = & \begin{bmatrix} a_1 & a_2  \end{bmatrix} M_{g_1}^{-1} \cdot M_{g_2}^{-1} \begin{bmatrix} a_1 \\ a_2  \end{bmatrix} > 0,\\ 
    &\forall g_1(x), g_2(x) \in \Sigma_g(x).
\end{split}
\end{equation}
For a two-link planar manipulator, the parameters $\xi_1, \xi_2, \xi_3$ are defined as $\xi_1  = I_1 + m_1l_{c_1}^2 + I_2 + m_2l_{c_2}^2 + m_2l_1^2$, $\xi_2 = I_2 + m_2l_{c_2}^2$, and $\xi_3 = m_2l_1l_{c_2}$,
where the notation is taken from \cite{JJS_book}.
\label{appdx:2}
\begin{pf} 
The verification of the positive definiteness can be completed by verifying the determinants associated with all upper-left submatrices of the Hermitian are positive. The first determinant of the upper-left submatrix:
\begin{equation*}
\begin{aligned}
    &(\xi_{1, g_1} + 2 \xi_{3, g_1}\cos(q_2))(\xi_{1, g_2} + 2 \xi_{3, g_2}\cos(q_2)) + \dots \\
    &\dots + (\xi_{2, g_1} + \xi_{3, g_1}\cos(q_2))(\xi_{2, g_2} + \xi_{3, g_2}\cos(q_2)) > 0,
\end{aligned}
\end{equation*}
can always be made true during the design of the manipulator by ensuring $\xi_1 > 2\xi_2$ and $\xi_2 > \xi_3$. To verify the second upper-left determinant, we leverage the simplifications above, and further express the parameters of one family in terms of another (\textit{e.g.} $m_{1, g_1(x)} = m_{1, g_2(x)} + \delta m_1$, assuming that all ``$\delta$'' terms take the form $\delta m = \frac{m}{K_m}$ where $K_m \not = 0$. We conduct the term-wise selection of the value of $q_2$ in order to find the lower bound of the determinant, which simplifies to:
\begin{equation*} \label{ul_derivative2}
\begin{aligned}
    &2l_1^4l_{c_2}^4m_1m_2^3 + l_1^4l_{c_2}^4m_1^2m_2^2 + \frac{l_1^4l_{c_2}^4m_1^2m_2^2}{K_1} - \frac{l_1^4l_{c_2}^4m_1^2m_2^2}{4K_1^2}- \\
    &\frac{l_1^5l_{c_2}^3m_1^2m_2^2}{4K_1^2} - \frac{l_1^6l_{c_2}^2m_1^2m_2^2)}{16K_1^2} + \frac{l_1^4l_{c_2}^4m_1^2m_2^2}{K_2} - \frac{l_1^4l_{c_2}^4m_1^2m_2^2}{4K_2^2}-\\
    &\frac{l_1^5l_{c_2}^3m_1^2m_2^2}{4K_2^2} - \frac{l_1^6l_{c_2}^2m_1^2m_2^2}{16K_2^2} + \frac{l_1^4l_{c_2}^4m_1m_2^3}{K_1} + \frac{3l_1^4l_{c_2}^4m_1m_2^3}{K_2} + \\
    &\frac{l_1^4l_{c_2}^4m_1m_2^3}{K_2^2} + \frac{l_1^4l_{c_2}^4m_1m_2^3}{K_1K_2} + \frac{3l_1^4l_{c_2}^4m_1^2m_2^2}{2K_1K_2} + \frac{l_1^5l_{c_2}^3m_1^2m_2^2}{2K_1K_2} + \\
    &\frac{(l_1^6l_{c_2}^2m_1^2m_2^2)}{8K_1K_2} > 0,
    \end{aligned}
\end{equation*}
where $K_1$ and $K_2$ reflect the uncertainty in the mass of link one and mass of link two respectively. $K_1$ and $K_2$ should be selected such that this expression is positive. In the closed interval, $K_1$ and $K_2$ should capture at least the maximum deviation between any two elements in the interval. Note, here that $K_1$ and $K_2$ can also take on negative values. The selection is dependent on the values of the dynamic parameters and the values of the ``$\delta$'' terms reflect the uncertainty in each parameter. If a parameter is known, the corresponding value of $K\rightarrow \pm\infty$. As the interval of uncertainty shrinks on both parameters, \textit{i.e.} $K_1\rightarrow \pm\infty$ and $K_2 \rightarrow \pm\infty$, implying that we have a small range of uncertainty on both parameters, we see that the above expression must be positive. 
\end{pf}

\section{Adaptive Control Law}\label{appendix: adaptive control law}
We use the adaptive control law defined in \cite{SL_OACRM} as our nominal control and estimation mechanism. The adaptive control law is characterized by the parameters: $K_D = 5*I, \Lambda = I$, $\Gamma = [60, 0, 0; 100, 0, 0; 0, 0, 20]$, and takes the form:
\begin{align} \label{slotine_aclaw}
    \tau_r &= \hat{M}(\theta)\ddot{\theta}_r + \hat{C}(\theta, \dot{\theta})\dot{\theta}_r - K_D s, \nonumber \\
    \dot{\hat{\xi}} &= \Gamma^{-1}Y^T(\theta, \dot{\theta}, \dot{\theta}_r, \ddot{\theta}_r),
\end{align}
where $\dot{\theta}_r = \dot{\theta_d} - \Lambda\tilde{\theta}$, $\ddot{\theta}_r = \ddot{\theta_d} - \Lambda\dot{\tilde{\theta}}$, $\tilde{\theta}(t) = \theta(t)-\theta_d(t)$, $s = \dot{\theta} - \dot{\theta}_r$, and $Y$ is a rearrangement of the linear parameterization described in \eqref{dyn1}. Here, $\theta_d(t)$ is the desired trajectory and $\tau_r$ is the reference control $u_r$. Additionally, as in \cite{SL_OACRM}, should the parameter estimate leave the given parameter range, we set the parameter to the boundary value until a subsequent estimation estimates a value in the interior of the interval.

\section{Proof of Theorem 1}\label{the1}
\begin{pf}
By \Cref{thm:esi}, conditions 3) and 4) imply that $\phi+\phi_\alpha$ meets the applicability conditions in \Cref{def: sufficient}. Thus we only need to show that the control generated by \Cref{alg:RSSA} always satisfies the condition $\dot{\phi}+\dot{\phi}_\alpha \leq -\eta$ when $\phi+\phi_\alpha \geq 0$ considering the ground truth dynamics. By \Cref{lemma: existence of robust safe control}, condition 2) implies that we can always choose a control in $\bar{\mathcal U}_S$ in \eqref{eq: robust safe set in g}, which satisfies $\dot\phi+\dot\phi_\alpha\leq -\eta$ when $\phi+\phi_\alpha\geq 0$ for any possible $g(x)\in\Sigma_g(x)$. Condition 1) says that the ground truth $g(x)$ corresponding to $\xi_T$ lies in $\Sigma_g(x)$. Hence the chosen control is safe with respect to the ground truth dynamics. Then \Cref{lemma: minimum effort} verifies that the control chosen in \Cref{alg:RSSA} is minimum effort. Hence \Cref{alg:RSSA} ensures the forward invariance of the safe set with minimum effort control.
\end{pf}

\end{document}